\pdfoutput=1

\documentclass[11pt]{article}

\usepackage[final]{acl}
\usepackage[]{acl}

\usepackage{times}
\usepackage{latexsym}

\usepackage[T1]{fontenc}

\usepackage[utf8]{inputenc}
\usepackage{microtype}

\usepackage{hyperref}
\usepackage{amsmath} 
\usepackage{amssymb} 
\usepackage{graphicx} 
 \usepackage{xcolor}
\usepackage{threeparttable}
\usepackage{pifont}
\usepackage{multirow}
\usepackage{subfigure}
\usepackage{graphicx}
\usepackage{booktabs}
\usepackage{enumitem}
\usepackage{algorithm}
\usepackage{algpseudocode}
\usepackage{amsmath}
\usepackage{graphicx} 
\usepackage{mdframed}
\usepackage{relsize}
\usepackage{ulem}
\usepackage{booktabs}
\usepackage{graphicx}
\usepackage{array}

\usepackage{inconsolata}
\NewDocumentCommand{\an}{ mO{} }{\textcolor{red}
{\textsuperscript{\textit{an}}\textsf{\textbf{\small[#1]}}}}
\NewDocumentCommand{\andong}{ mO{} }{\textcolor{blue}
{\textsuperscript{\textit{andong}}\textsf{\textbf{\small[#1]}}}}

 \usepackage{amsmath}

\usepackage{csquotes}
\usepackage{epigraph}
\usepackage{multirow} 
\usepackage{geometry}
\usepackage{array}
\usepackage{longtable}
\usepackage{float}
\usepackage{mdframed}
\usepackage{xcolor}
\definecolor{deepyellow}{rgb}{0.8, 0.7, 0.0} 
\definecolor{deepgreen}{rgb}{0.0, 0.7, 0.0} 

\definecolor{chatgpt_c}{RGB}{121,147,210}
\definecolor{src_c}{RGB}{238,154,189}
\definecolor{tgt_c}{RGB}{139,212,209}
\definecolor{src_tgt_c}{RGB}{149,149,149}
\definecolor{ibut_c}{RGB}{247,182,95}

\definecolor{chinese_red}{RGB}{230,239,255}
\definecolor{chinese_red_small}{RGB}{252,255,230}
\definecolor{chinese_brown}{RGB}{246,230,255}
 \usepackage{booktabs}

\definecolor{win}{RGB}{165,127,183} 
\definecolor{tie}{RGB}{204,161,189}
\definecolor{loss}{RGB}{229,207,221}

\usepackage{CJK}

\title{Large Language Models for Classical Chinese Poetry Translation: \\Benchmarking, Evaluating, and Improving}

\author{
Andong Chen\thanks{Work was done when Andong Chen was at Pengcheng Laboratory.}$^1$\hspace{0.5mm}, 
 Lianzhang Lou$^{2}$\hspace{0.5mm}, 
\textbf{Kehai Chen}\thanks{~~Corresponding author.}$^{1,2}$\hspace{0.5mm}, 
 \textbf{Xuefeng Bai}$^{1}$\hspace{0.5mm}, 
 \textbf{Yang Xiang}$^{2}$\hspace{0.5mm}, \\
 \textbf{Muyun Yang}$^{1}$\hspace{0.5mm},
 \textbf{Tiejun Zhao}$^{1}$\hspace{0.5mm}, 
 \textbf{Min Zhang}$^{1}$\hspace{0.2mm}\hspace{1.5mm} \\
 $^1$ School of Computer Science and Technology, Harbin Institute of Technology, China\\
 $^2$ Pengcheng Laboratory, Shenzhen, China \\
  ands691119@gmail.com,  \{loulzh, xiangy\}@pcl.ac.cn  \\
  \{chenkehai, baixuefeng, yangmuyun, tjzhao, zhangmin2021\}@hit.edu.cn, 
}

\begin{document}

\maketitle
\begin{abstract}
Different from the traditional translation tasks,  classical Chinese poetry translation requires both adequacy and fluency in translating culturally and historically significant content and linguistic poetic elegance.
Large language models (LLMs) with impressive multilingual capabilities may bring a ray of hope to achieve this extreme translation demand. 
This paper first introduces a suitable benchmark (\textbf{PoetMT}) where each Chinese poetry has a recognized elegant translation.
Meanwhile, we propose a new metric based on GPT-4 to evaluate the extent to which current LLMs can meet these demands.
Our empirical evaluation reveals that the existing LLMs fall short in the challenging task. 
Hence, we propose a \textbf{R}etrieval-\textbf{A}ugmented machine \textbf{T}ranslation (RAT) method which incorporates knowledge related to classical poetry for advancing the translation of Chinese Poetry in LLMs.
Experimental results show that RAT consistently outperforms all comparison methods regarding wildly used BLEU, COMET, BLEURT, our proposed metric, and human evaluation.

\end{abstract}

%

\section{Introduction}

\begin{displayquote}
  \textit{The three difficulties in translation are: adequate, fluent, and elegant.}
  \begin{flushright}
  -- \citealp{tianyanlun}
  \end{flushright}
\end{displayquote}

The emergence of large language models (LLMs), especially ChatGPT, has demonstrated impressive performance in translation tasks \cite{tyen2023llms,liang2023encouraging,DBLP:journals/corr/abs-2303-16104,DBLP:journals/corr/abs-2308-14186,zhao2024review,zhang-etal-2024-paying,chen-etal-2024-dual}. As the requirements for translation quality continues to rise, translated results not only be adequate but also fluent and elegant \cite{Wang2024WhatIT,huang2024aligning,gao2024machine,wu2024perhaps}. This raises a question: can existing LLMs meet such translation requirements, and if so, to what extent can they achieve this performance?


To answer this question, we introduced a suitable benchmark (PoetMT): translating classical Chinese poetry into English. Firstly, these poems carry culture and history, so the translated results need to adequately convey these meanings. Secondly, classical Chinese poetry has strict rules on rhyme, tone, and structure, making fluent translation a significant challenge. Lastly, classical Chinese poetry has aesthetic value, with the concise expressions of the classical Chinese language showing linguistic poetic elegance, which needs to be preserved in translated results.

Based on the proposed PoetMT benchmark, previous automatic evaluation metrics for machine translation only analyze entire sentences without evaluating classical poetry translation quality explicitly \cite{papineni2002bleu,rei2022comet,sellam-etal-2020-bleurt}.
To overcome the limitations of traditional evaluation metrics, we propose an automatic evaluation metric based on GPT-4 (OpenAI 2024), which better evaluates translation quality from the perspectives of adequacy, fluency, and elegance. Additionally, evaluating current LLM-based MT methods reveals that these translated results often lack historical and cultural knowledge, strict rhyme and structure rules, and concise expressions. To address these issues, we introduce RAT, a retrieval-augmented machine translation method. This method enhances the translation process by retrieving knowledge related to classical poetry, ensuring the translated output remains adequate to the original text while also achieving fluency and elegance.

To our knowledge, this is the first study evaluating the translation performance of LLMs based on the task of translating classical Chinese poetry. Through this effort, we aim not only to test the capabilities of LLMs in translating classical Chinese poetry but also to inspire community discussion on the potential and future development of LLMs in translated texts that are adequate, fluent, and elegant.

Our contributions are summarized as follows:
\begin{itemize}
    \item  We have introduced the first classical poetry translation benchmark  (PoetMT), which allows for a better evaluation of LLMs in terms of adequacy, fluency, and elegance. 
    \item We have designed a new evaluation metric based on GPT-4 to evaluate classical poetry translation. This metric aligns more closely with human annotations and is better suited for the PoetMT benchmark.
    \item Based on the limitations of current LLM-based translation methods on the PoetMT benchmark, we have proposed a retrieval-augmented translation method to enhance the performance of LLMs in this task.
\end{itemize}

\section{Related Work}

\textbf{Poetry Machine Translation} The initial attempts at poetry machine translation utilized phrase-based machine translation systems to translate French poetry into metrical English poetry \cite{genzel2010poetic}. This study demonstrated how statistical machine translation systems can produce translations while adhering to the rhythmic and rhyming rules of poetry. \citet{chakrabarty-etal-2021-dont} conducted an empirical study that highlighted a critical but often overlooked issue: even though they can maintain meaning and fluency, advanced machine translation systems trained on large amounts of non-poetry data struggle to preserve the original style of poetry. To address the issue of style preservation, some studies have managed to generate diverse styles by inputting sentences with specific styles into the encoder and embedding the target style into the decoder \cite{DBLP:journals/corr/abs-1803-01557,liu2012sentence}. Considering that poetry often contains richly stylized semantic information and deep historical and cultural connotations, such as in classical Chinese poetry, \citet{doi:10.1080/0952813X.2020.1836033} proposed a Hybrid Machine Translation model to improve the semantic and syntactic accuracy of poetry translations. Recently, \citet{Wang2024WhatIT} have successfully implemented translations of modern poems from English to Chinese using the knowledge and multilingual capabilities of ChatGPT.

\textbf{Chinese Classical Poetry Dataset} Chinese classical poetry has several open datasets. \citet{DBLP:conf/ijcai/ChenYSLYG19} published the first fine-grained dataset, annotating a manually fine-grained emotional poetry corpus containing 5,000 Chinese quatrains. \citet{2020-6-1252} released a dataset containing 3,940 quatrains with automatically annotated themes and 1,917 quatrains with automatically annotated emotions, using a template-based method. Additionally, \citet{DBLP:journals/talip/LiuYQL20} collected a parallel bilingual dataset of ancient and modern Chinese and aligned the lines using a string-matching algorithm. Based on these bilingual pairs, \citet{DBLP:journals/corr/abs-2106-01979} constructed a matching dataset to evaluate models' semantic understanding capabilities. To our knowledge, this is the first benchmark for English translation of Chinese classical poetry, aimed at evaluating the translation performance of current large models in terms of ``adequate, fluent and elegant".

\section{Classical Chinese Poetry Dataset Construction}
In this section, we discuss the design and construction of the PoetMT benchmark, including the rules and steps for building this benchmark.

\subsection{Discourse-Level Poetry Translation}
\begin{figure}[ht] 
\centering
\includegraphics[width=0.25\textwidth]{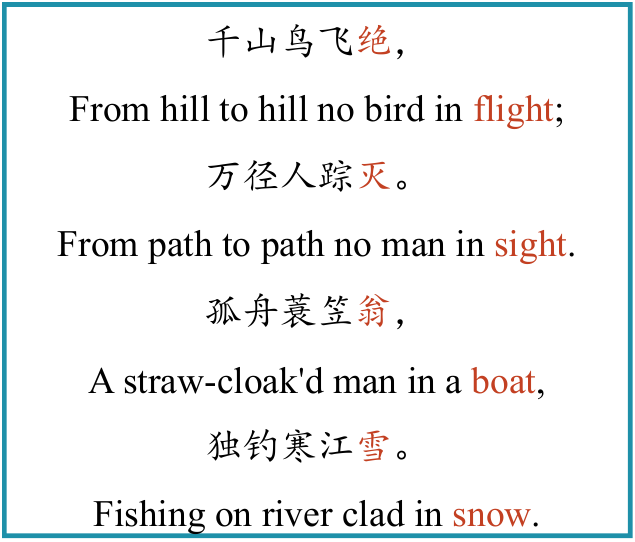} 
\caption{An example block in the fluency and elegance in discourse-level poetry translation. The \textcolor{red}{red parts} indicate rhymes in both English and Chinese.} 
\label{doucment}
\end{figure}

\begin{table*}[ht]\small \centering
\begin{tabular}{l|cccc}
\hline
Poem Type & \multicolumn{1}{l}{Number of Poem} & \multicolumn{1}{l}{Unique Token} & \multicolumn{1}{l}{Average Tokens Per Sentence} & \multicolumn{1}{l}{Total Token Numbers} \\ \hline
Tang      & 197                                & 1980/3839                        & 11.7/13.4                                       & 11727/13115                             \\
Song      & 189                                & 2214/4899                        & 10.9/14.1                                       & 16984/18212                             \\
Yuan      & 222                                & 2006/3650                        & 12.8/13.2                                       & 12145/1197                              \\
Total     & 608                                & 3059/9223                        & 11.69/13.6                                      & 40856/42524                             \\ \hline
\end{tabular}
\caption{Statistics on the benchmark. Numbers a/b denote the corresponding number in source/target sentences.}
\label{stat}
\end{table*}

We collected a batch of classical Chinese poetry data and corresponding human English translations from online resources\footnote{We selected professional translation versions available online, specifically those authored by Xu Yuanchong. Therefore, the translation results are from very experienced poem translators.} The classical Chinese poetry database contains 1200 poems from Tang Poems, Song Poems, and Yuan Opera. Then, We manually screened 608 classical Chinese poems and their corresponding translations to serve as gold standards for translation source and target alignment\footnote{The manually filtered content primarily excludes poems with multiple translation versions, misaligned or incomplete Chinese and English poems, and poems with factual errors.}. As shown in Figure \ref{doucment}.

The statistics of the PoetMT benchmark are shown in Table \ref{stat}. We present the number of classical Chinese poetry, the number of unique tokens, the average number of tokens per sentence, and the total number of tokens in different poetry types. The source sentences in this benchmark have a moderate length, and the selected target translation sentences are well-aligned with the source in terms of length, indirectly reflecting the high quality of the reference sentences.

\subsection{Adequacy in Sentence-Level Translation}

\begin{figure}[h] 
\centering
\includegraphics[width=0.24\textwidth]{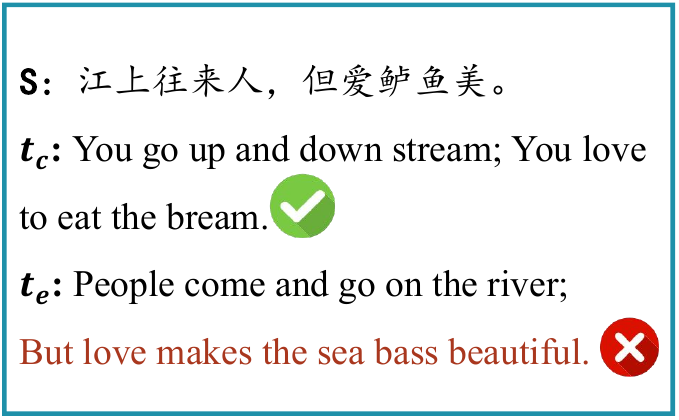} 
\caption{An example block in the adequacy in sentence-level poetry translation.} 
\label{adequacy example}
\end{figure}

Due to the inclusion of historical background and common knowledge in classical Chinese poetry, achieving adequacy in translation poses a significant challenge. Therefore, to conduct a more detailed evaluation of adequacy, we have constructed a sentence-level test set.

Following related works\cite{DBLP:conf/emnlp/HeWXL20,yao-etal-2024-benchmarking}, we selected sentences containing historical knowledge and commonsense from the collected data of classical Chinese poetry. We avoid selecting semantically similar words to ensure diversity in the test set. Additionally, we preferred to select words that have different English translations depending on the context. This test set consists of 758 sentences, with each translated sentence represented as a triplet ($s$, $t_c$, $t_e$). As shown in Figure \ref{adequacy example}, $s$ contains the source sentence with ambiguous words, $t_c$ and $t_e$ are the contrast translations, with the former being correct and the latter incorrect. 


\subsection{Classical Chinese Poetry Knowledge Base}
\label{RAT}

Classical Chinese poetry holds rich historical and cultural nuances, but due to the limited resources for Classical Chinese, modern Chinese knowledge can greatly mitigate this issue. The PoetMT benchmark includes a Classical Chinese Poetry Knowledge Base collected from open-source projects and internet resources. This Knowledge Base consists of 30,000 entries, including 30,000 Classical Chinese poems along with knowledge such as their corresponding historical background, dynasty name, modern Chinese translation, author introduction, modern Chinese analysis, and poetry type. The case is displayed in Appendix \ref{base_case_section}.

\section{LLM-based Evaluation Method}
\label{llm-eval}
\subsection{Evaluation Criteria}

\begin{figure}[ht] 
\centering
\includegraphics[width=0.45\textwidth]{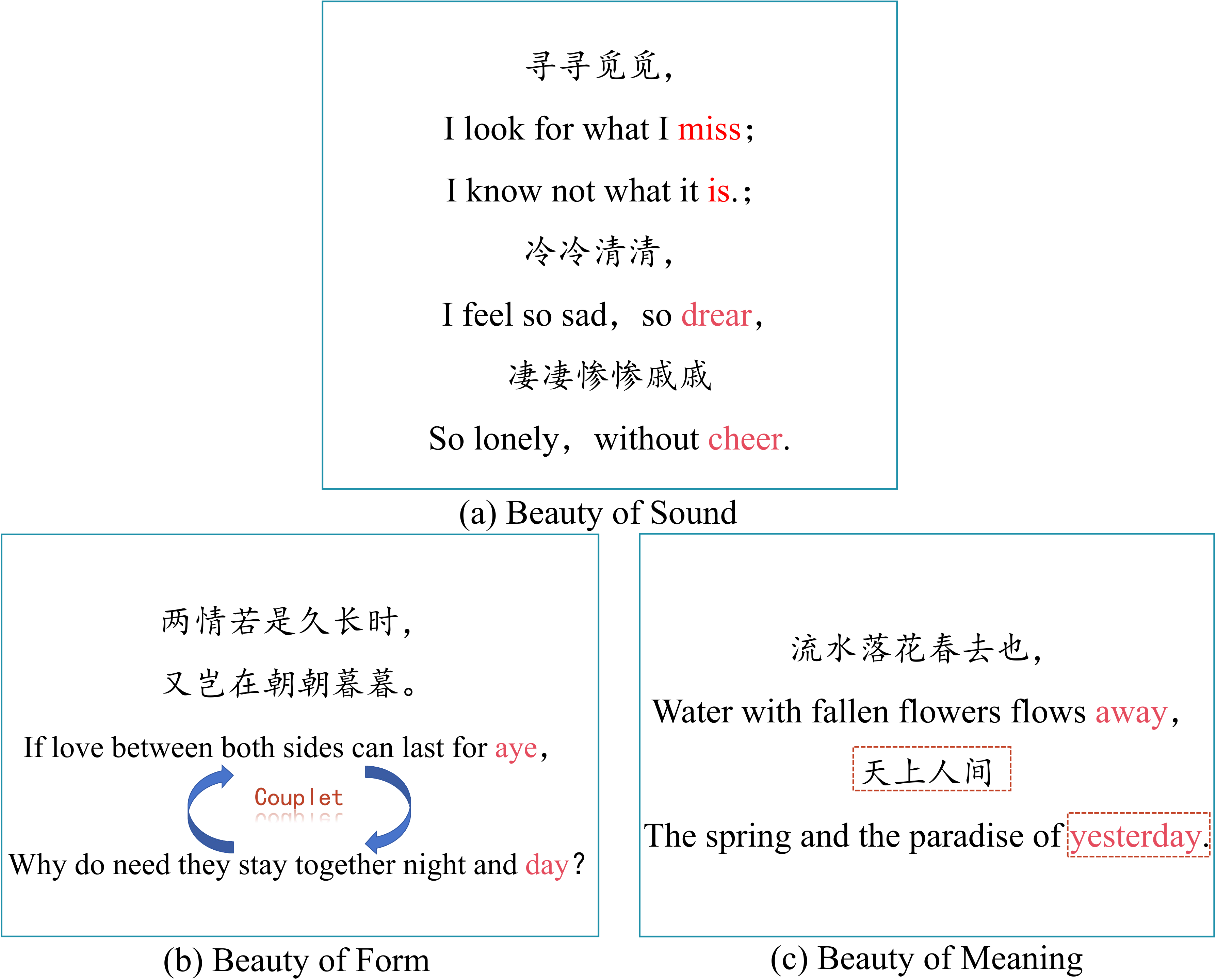} 
\caption{Examples of different evaluation metrics. Figure (a) represents the rhyme of the final words; Figure (b) shows that the two translated sentences have the same word count and couplet structure; Figure (c) indicates the accurate translation capturing the implied meaning of time passing.} 
\label{metric_examples}
\end{figure} 
\vspace{-0.3cm}

The translation of classical poetry requires not only artistic expression but also an understanding of the cultural background, yet the premise of correctness does not imply a singular or unique expression.
Following this line of thought, we evaluate the quality of classical poetry translations through three aspects: ``Adequate," ``Fluent," and ``Elegant."

\subsubsection{Adequate Criteria}
\textbf{Accuracy (Acc)↑}: Focus on the precision of each element in the translation, accurately translating historical, cultural, and factual aspects, including words and phrases, to maintain the correct semantic and logical relationships of the poem.

\subsubsection{Fluent Criteria}
\textbf{Beauty of Sound (BS)↑}: The beauty of sound in Chinese classical poetry is primarily reflected in its rhyme. This standard examines whether the translation achieves harmonious sound, adherence to strict metrical rules, and a rhythm that is both smooth and dynamic. As shown in Figure \ref{metric_examples}(a).

\textbf{Beauty of Form (BF)↑}: Chinese classical poetry emphasizes symmetrical structures, with common forms including the "Five-character eight-line regulated verse (wulü)", "Seven-character eight-line regulated verse (qilü)", and "Extended forms (pailü)" among others. Each form showcases the structural characteristics of Chinese poetry. This standard evaluates whether the translation maintains consistency with the source poem's structure, including the alignment of line numbers and balanced phrasing. As shown in Figure \ref{metric_examples}(b).

\subsection{Elegant Criteria}
\textbf{Beauty of Meaning (BM)↑}: Chinese classical poetry uses concise and precise language to create vivid imagery and a rich atmosphere for readers. The criteria evaluate the depth and richness of the translation, focusing on the effectiveness of conveying themes, emotions, and messages. As shown in Figure \ref{metric_examples}(c).

\subsection{LLM-based Classical Poetry Metric}

In this section, we aim to establish a method for evaluating the translation of classical Chinese poetry based on LLMs. Drawing from current QE (Quality Estimation) research \cite{Li2023TranslateMN,kocmi-federmann-2023-large-v1}, we have developed an approach to utilize LLM in determining the quality of classical Chinese poetry translations. We designed a 1-5 scoring prompt to help LLM focus on the translation quality in terms of Beauty of Sound (LLM-BS). In the evaluation criteria, a score of 1 reflects a poor translation of classical Chinese poetry, a score of 3 indicates a basic but imperfect translation result and a score of 5 represents an excellent and accurate translation in this aspect. Subsequently, test data is provided to the LLM, and the LLM is asked to generate an evaluation with scores only. Additionally, we calculate the average scores (LLM-Avg) of BF, BF, and BM to evaluate the overall translation performance. The prompt details for Beauty of Sound(LLM-BS), Beauty of Form (LLM-BF) and Beauty of Meaning (LLM-BM) are provided in Appendix \ref{bs_score_appendix}, \ref{bf_score} and \ref{bm_score}.

\section{Proposed Method: RAT}

\begin{figure*}[!h] 
\centering
\includegraphics[width=0.85\textwidth]{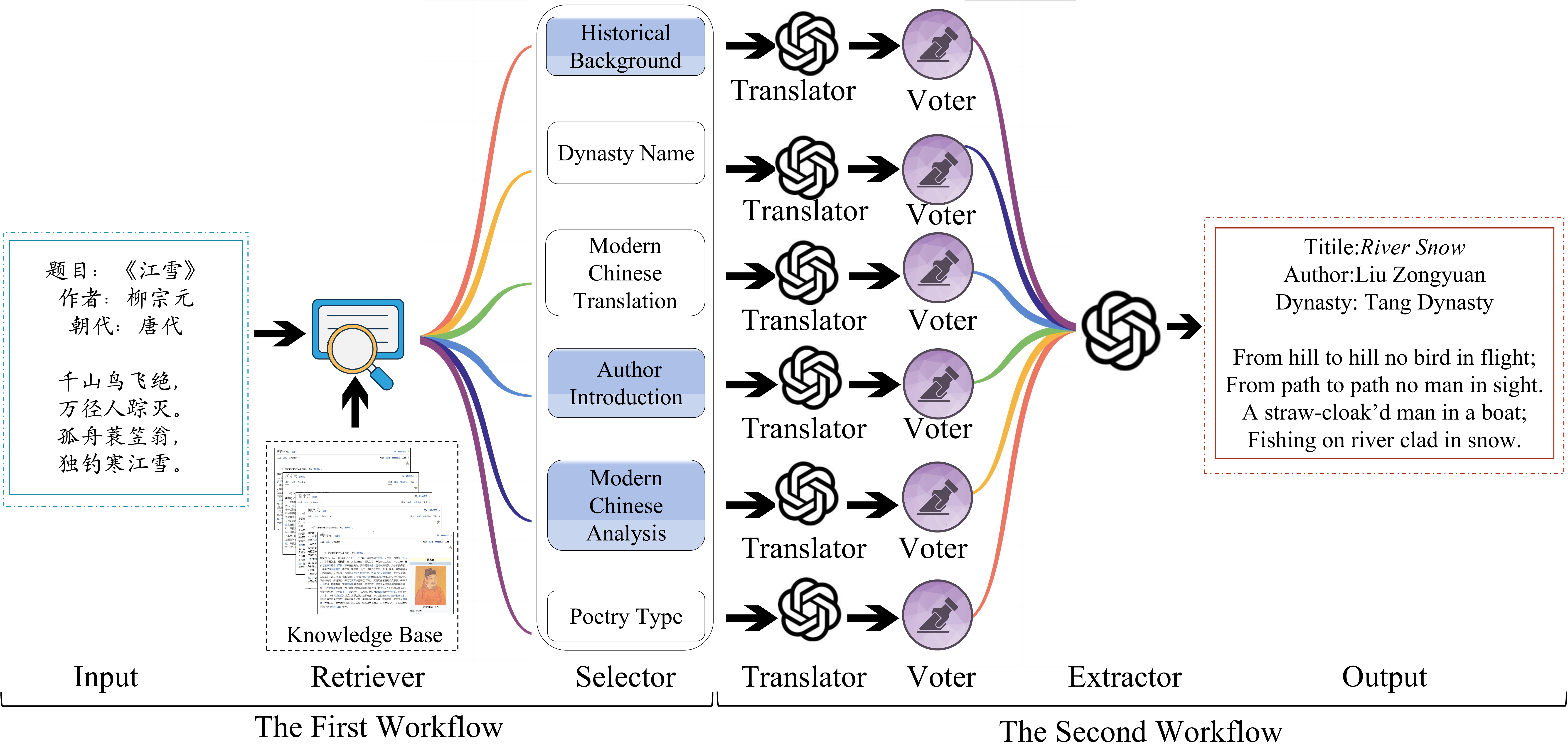} 
\caption{The proposed RAT framework. The "Historical Background," "Author Introduction," and "Modern Chinese Analysis" parts are at the discourse level, so the Selector needs to make selections based on the content.} 
\label{framework}
\end{figure*}

The RAT (\textbf{R}etrieval \textbf{A}ugmented \textbf{T}ranslate) method was introduced to enhance translation performance by leveraging rich contextual information from the Classical Chinese Poetry Knowledge Base. We introduce the two workflows implemented in the RAT method in detail (in Figure \ref{framework}). The first workflow uses poetry as input and retrieves relevant knowledge about the poetry from the Classical Chinese Poetry Knowledge Base through text-matching methods, obtaining different views of the relevant knowledge. The second workflow translates based on the relevant knowledge from different views and integrates these to produce the final translation result.

\subsection{The First Workflow}

In the first workflow of RAT, there are two modules: Retriever and Selector. 

\textbf{Retriever.} We propose a retrieval augmentation method to obtain knowledge relevant to translating classical Chinese poetry. Based on the Classical Chinese Poetry Knowledge Base, we use text-matching methods to retrieve uniquely relevant knowledge from multiple perspectives\footnote{The Classical Chinese Poetry Knowledge Base contains all 608 classical poems presented in the paper, ensuring a one-to-one correspondence between the poems and the knowledge.}. These perspectives include historical background, dynasty name, modern Chinese translation, author introduction, modern Chinese analysis, and type. 

\textbf{Selector.} The goal of the selector is to filter out irrelevant content to translated sentences from the results of the retriever. As an agent of the LLM, the selector first understands the historical background, author introduction, and modern Chinese analysis based on the source sentences, and then outputs content relevant to the translated source sentences. Specific prompts are displayed in Appendix \ref{detail stage1}.

\subsection{The Second Workflow}

In the second workflow of RAT, there are three modules: Translator, Voter, and Extractor. 

\textbf{Translator.} The goal of the Translator is to translate classical Chinese poetry based on different types of retrieved knowledge. Six types of related knowledge were retrieved for classical Chinese poetry, resulting in six different translation outputs.  Specific prompts are displayed in Appendix \ref{detail stage2}.

\textbf{Voter.} The purpose of the Voter is to integrate translations based on different retrieval results to improve translation quality. As an agent of the LLM, the Voter votes on each sentence of all translation outputs based on the source input selects the highest-quality translations and concatenates these selected sentences to form the final translation result.  Specific prompts are displayed in Appendix \ref{detail stage3}.

\textbf{Extractor.} The goal of the Extractor is to extract the final translation results generated by the Voter. As an agent of the LLM, the Extractor filters out noise from the content generated by the Voter based on the source input and outputs the final translation results. Specific prompts are displayed in Appendix \ref{detail stage4}.


\section{Experiment Setup}

\subsection{Comparing Systems}
In our evaluation, RAT is compared with a range of translation methods, including Zero-shot \cite{DBLP:conf/iclr/WeiBZGYLDDL22}, 5-shot \cite{hendy2023good}, Rerank \cite{DBLP:conf/eamt/MoslemHKW23}, Refine \cite{chen2023iterative}, MAD \cite{liang2023encouraging}, EAPMT \cite{Wang2024WhatIT}, and Dual-Reflect \cite{chen-etal-2024-dual}. To validate its generalizability, we utilize three LLMs, which include closed-source models such as ChatGPT\cite{ouyang2022training} and GPT-4\cite{achiam2023gpt} \footnote{The ChatGPT and GPT-4 models used in this work are accessed through the gpt-3.5-turbo and gpt-4 APIs, respectively.}, as well as open-source models like Llama3-8B \cite{dubey2024llama} \footnote{https://huggingface.co/meta-llama/Meta-Llama-3-8B} and Vicuna-7B \cite{vicuna2023} \footnote{https://huggingface.co/lmsys/vicuna-7b-v1.5}.  Details on comparative methods are in Appendix \ref{sec:comparative_methods}. 
\vspace{-0.3cm}
 
\subsection{Evaluation Metrics}
\textbf{LLM-based Automatic Evaluation.} We propose an automatic evaluation method for translation based on LLMs in the section \ref{llm-eval}.

\textbf{Traditional Automatic Evaluation.} We initially employ COMET\footnote{https://huggingface.co/Unbabel/wmt22-comet-da} \cite{rei2022comet} and BLEURT\footnote{https://github.com/lucadiliello/bleurt-pytorch} \cite{sellam2020bleurt} as automatic metrics, aligning with the established standards in LLM-based translation literature \cite{He2023ExploringHT,huang2024aligning}. For traditional translation evaluation, we use BLEU \footnote{https://github.com/mjpost/sacrebleu} \cite{papineni2002bleu}. 

\section{Experimental Results}

\begin{table*}[!h]\centering
\scalebox{0.650}{
\begin{tabular}{lcccccccccc}
\hline
\multirow{2}{*}{Methods} & \multicolumn{10}{c}{Discourse-Level Poetry Translation}                                                                                                                                                                                                                            \\ \cline{2-11} 
                         & \multicolumn{1}{l}{COMET $\uparrow$} & BLEURT $\uparrow$ & LLM-BM $\uparrow$ & \multicolumn{1}{l}{LLM-BS $\uparrow$} & \multicolumn{1}{l}{LLM-BF $\uparrow$} & \multicolumn{1}{l}{LLM-Avg $\uparrow$} & \multicolumn{1}{l}{BLEU-1 $\uparrow$} & \multicolumn{1}{l}{BLEU-2 $\uparrow$} & \multicolumn{1}{l}{BLEU-3 $\uparrow$} & \multicolumn{1}{l}{BLEU-4 $\uparrow$} \\ \hline
\textbf{GPT-4}                    & 60.3                      & 43.0   & 4.0      & 3.7                          & 3.6                            & 3.8                           & 22.1                       & 7.8                        & 3.3                        & 1.7                        \\
\textbf{ChatGPT}                  & 61.1                      & 42.4   & 3.3      & 3.2                          & 2.9                          & 3.1                          & 23.4                       & 8.7                        & 3.1                        & 1.8                        \\
\quad+5shot                   & 61.0                      & 42.5   & 3.5      & 3.3                          & 3.3                          & 3.4                           & 22.0                       & 7.7                        & 3.2                        & 1.6                        \\
\quad+Rerank                  & 61.0                      & 42.5   & 3.7      & 3.7                          & 3.9                          & 3.8                           & 22.5                       & 8.0                        & 3.4                        & 1.7                        \\
\quad+MAD                     & 59.9                      & 42.3    & 3.7      & 3.6                          & 3.8                          & 3.7                           & 23.2                       & 8.8                        & 3.7                        & 1.8                        \\
\quad+Dual-Reflect         & 58.2                      & 40.9   & 3.8      & 3.8                          & 3.9                          & 3.8                           & 20.5                       & 7.5                        & 3.2                        & 1.6                        \\
\quad+EAPMT                   & 61.1                      & 42.9   & 3.8         & 3.7                          & 3.8                          & 3.7                           & 21.6                       & 7.5                        & 3.1                        & 1.5                        \\
\quad+RAT                     & \textbf{62.7}             & \textbf{43.9} & \textbf{4.1}      & \textbf{3.9}                          & \textbf{3.9}                          & \textbf{4.0}                          & \textbf{23.9}              & \textbf{9.8}               & \textbf{3.9}               & \textbf{2.2}               
\\ \hline

\textbf{Vicuna-7B}                & 52.2                    & 26.4   & 2.4      & 2.4                          & 1.8                          & 2.2                           & 16.5                          & 4.7                          & 3.4                          & 1.0                          \\
\quad+5shot                   & 52.4                      &26.1   & 2.5      & 2.6                          &  2.3                         & 2.4                           & 17.1                           & 4.3                          & 3.6                          & 1.3                          \\
\quad+Rerank                  & 52.8                      & 26.3   & 3.0      & 2.6                          & 3.3                          & 2.8                           & 17.5                          & 5.0                          & 3.7                          & 1.6                          \\
\quad+RAT                     & \textbf{60.1}                     & \textbf{26.9}   & \textbf{3.0}      & \textbf{2.5 }                         & \textbf{3.3 }                        & \textbf{2.9}                           & \textbf{17.6}                          & \textbf{5.3 }                         & \textbf{3.9}                          & \textbf{1.9}                          \\ \hline
\textbf{Llama3-8B}                & 54.3                      & 37.4   & 2.7      & 2.6                          & 2.4                          & 2.5                           & 17.4                          &6.1                          & 3.5                          & 1.3                          \\
\quad+5shot                   & 54.5                      & 37.6   & 2.9      & 2.8                          & 2.6                          & 2.7                           & 17.4                          &   6.2                        & 3.4                          &  1.3                         \\
\quad+Rerank                  & 54.8                      & 38.1   & 3.0      & 3.3                          & 3.5                          & 3.2                           & 17.9                          & 6.6                          & 3.6                          & 1.5                          \\
\quad+RAT                     & \textbf{55.6 }                     & \textbf{38.4}   & \textbf{3.4}     & \textbf{3.3}                         & \textbf{3.6}                          & \textbf{3.4}                           & \textbf{18.2}                          & \textbf{7.0}                          & \textbf{3.9}                          & 1.8                          \\ \hline
\textbf{Qwen-72B}                & 60.9                      & 43.5   & 3.4      & 3.4                          & 3.3                          & 3.3                           & 22.1                         &           7.1               & \textbf{3.0}                          & 2.0                          \\
\quad+5shot                   & 60.4                      & 43.8   & 3.6      &  \textbf{3.7}                          & 3.4                          & 3.5                           & 21.5                          &   7.2                        & 2.9                          &  1.5                         \\
\quad+Rerank                  & 59.8                      & 43.2   & 3.0      & 3.3                          & 3.5                          & 3.2                           & 20.6                          & 6.7                          & 2.7                          & 1.3                          \\
\quad+RAT                     & \textbf{61.7 }                     & \textbf{43.5}   & \textbf{3.7}     & 3.6                         & \textbf{3.6}                          & \textbf{3.6}                           & \textbf{22.9}                          & \textbf{8.0}                          & 2.9                          & \textbf{2.0} 

\\ \hline
\end{tabular}
}
\caption{The main results from the PoetMT benchmark are presented. The bold indicates the highest scores. The bolded results indicate the highest statistically significant scores (p-value $< 0.05$ in the paired t-test against all compared methods).}
\label{main_results}
\end{table*}

\subsection{Can LLM evaluate Classical Poetry ?}
We first translate randomly selected 100 discourse-level data from the PoetMT benchmark by the RAT method. Inspired by \citealp{Li2023TranslateMN}, we then scored the translations according to the citeria outlined in Figures \ref{fig:poetry-evaluation-bs}, \ref{fig:poetry-evaluation_bf}, and \ref{fig:poetry-evaluation_bm}. We compare the different evaluation results of the different automatic methods with human-annotated results to calculate the Pearson correlation coefficient \cite{pearson1920notes}, Spearman correlation coefficient \cite{spearman1961proof}, and Kendall correlation coefficient \cite{kendall1948rank} to determine the level of consistency.


\begin{table}[!ht]\centering
\resizebox{0.75\linewidth}{!}{%
\begin{tabular}{@{}lccc@{}}
\toprule
Metric  & Pearson's $r$ $\uparrow$ & Spearman's $\rho$ $\uparrow$ & Kendall's $\tau$ $\uparrow$ \\ \midrule
        & \multicolumn{3}{c}{Traditional Automatic Evaluation} \\ \midrule
BLEU    & -0.23         & -0.18             & -0.12            \\
BLEU-1  & 0.05          & 0.08              & 0.05             \\
BLEURT  & 0.14          & 0.16              & 0.11             \\
COMET   & 0.13          & 0.18              & 0.11             \\ \midrule
        & \multicolumn{3}{c}{Qwen-72B-based Automatic Evaluation}                         \\ \midrule
LLM-BM  & 0.63          & 0.59              & 0.61             \\
LLM-BF  & 0.53          & 0.55              & 0.50             \\
LLM-BS  & 0.54          & 0.53              & 0.55             \\
LLM-AVG & 0.57          & 0.53              & 0.54             \\ \midrule
        & \multicolumn{3}{c}{GPT-4-based Automatic Evaluation}                           \\ \midrule
LLM-BM  & 0.85          & 0.81              & 0.85             \\
LLM-BF  & 0.71          & 0.75              & 0.70             \\
LLM-BS  & 0.73          & 0.73              & 0.76             \\
LLM-AVG & 0.77          & 0.73              & 0.75             \\ \bottomrule
\end{tabular}}
\caption{correlation metrics between human and BLEU, BLEU-1, COMET, BLEURT, LLM-BM, LLM-BF, LLM-BS or LLM-AVG evaluation on our PoetMT.}
\label{corr_score}
\end{table}


The results in Table \ref{corr_score} indicate that large language models can serve as effective tools for evaluating the translation quality of classical Chinese poetry. In contrast, BLEU, COMET, and BLEURT were found to be less consistent with human evaluations, highlighting the advantages of our proposed evaluation method in the translation of classical poetry.

\subsection{Main Results}
\label{section_main_results}

We compare various different LLM-based methods on the PoetMT benchmark with RAT. The results are shown in Table \ref{main_results}. The analysis of the experimental results is as follows: 


\textbf{The task of translating Classical Chinese Poetry is challenging.} The experiment shows that translating Classical Chinese Poetry is difficult, with low COMET/BLEURT/BLEU scores for current LLM-MT methods. This is because the translation is challenging and n-gram-based evaluation methods are not ideal for poetry. Additionally, GPT-4-based evaluation metrics, particularly in BS/BM/BF/AVG aspects, still need considerable improvement.

\textbf{The effectiveness of RAT method.} The proposed RAT method consistently outperforms other comparative methods across all evaluation metrics, demonstrating the effectiveness of our approach.
    
\textbf{Performance Variations Among Different Types of LLMs.} Among all comparative methods, closed-source models perform better on this task than open-source models, possibly implying that closed-source models benefit from richer pre-training data, thus enabling higher-quality translations. This also suggests that the PoetMT task is more challenging.
    
\textbf{The effectiveness of retrieved knowledge.} For methods that rely on LLMs' self-generated knowledge (such as the EAPMT method), the RAT approach, which utilizes retrieval-based knowledge, provides more accurate information and thus enhances translation quality. This suggests that more accurate relevant knowledge can better assist in the PoetMT task.



\subsection{Evaluation of Adequacy}


To evaluate the translation adequacy of LLMs, we used a dataset of 758 Classical Chinese sentence-level translations. Following \citealp{liang2023encouraging} and \citealp{chen2024dual}, we evaluated translations through manual adequacy assessment (Appendix \ref{sec:human_evaluation}), the LLM-BM score (GPT-4), and the human BM score (Appendix \ref{human_evaluation_bm}).

Results (Table \ref{human_main}) show that RAT achieved the best adequacy scores. This suggests that retrieving accurate information improves adequacy. RAT also received the highest LLM-BM score, confirming that it better captures the themes, emotions, and messages of the original poems.

\begin{table}[!ht]\centering
\scalebox{0.66}{ 
\begin{tabular}{@{}lccc@{}}
\toprule
Methods                            & LLM-BM $\uparrow$       & \multicolumn{1}{l}{Human-BM $\uparrow$} & ACC $\uparrow$           \\ \midrule
\textbf{GPT-4}                     & 3.9          & 3.6                                  & \textbf{69.1} \\ \midrule
\textbf{ChatGPT}                   &              &                                      &               \\
\quad +Zero-Shot    & 3.2          & 3.2                                  & 60.5          \\
\quad +Rerank       & 3.2          & 3.3                                  & 64.4          \\
\quad +Dual-Reflect & 3.7          & 3.6                                  & 66.4          \\
\quad +MAD          & 3.7          & 3.8                                  & 67.3          \\
\quad +RAT          & \textbf{3.9} & \textbf{3.9}                         & \textbf{69.9} \\ \midrule
\textbf{Vicuna-7B}                   &              &                                      &               \\
\quad +Zero-Shot    & 2.1          & 0.8                                  & 26.9          \\
\quad +Rerank       & 2.3          & 1.2                                  & 31.7         \\
\quad +Dual-Reflect & 2.0          & 1.0                                  & 33.0          \\
\quad +MAD          & 2.2          & 1.3                                  & 67.3          \\
\quad +RAT          & \textbf{2.5} & \textbf{2.1}                         & \textbf{43.4} \\

\bottomrule
\end{tabular}}
\caption{The LLM-BM and human-annotated results of the Adequacy in Sentence-Level PoetMT benchmark Translation. The results of Llama3-8B and Qwen-72B are in Appendix \ref{adequcay_of_llm}.}
\label{human_main}
\end{table}

\subsection{Data Validation Experiments}


To explore whether PoetMT poems were included in the training data of closed-source LLMs like GPT-4 and ChatGPT (\textsection \ref{section_main_results}), we conducted an experiment using 150 poems (50 each from Tang poetry, Song lyrics, and Yuan opera). Following concerns raised by \cite{DBLP:conf/iclr/ShiAXHLB0Z24}, we prompted GPT-4/ChatGPT with the title and author to generate poems, then evaluated the similarity to human reference using SacreBLEU. As shown in Table \ref{data_validation}, the results indicate higher BLEU scores for Chinese and lower scores for English, suggesting fewer English translations in LLM training data. 

\begin{table}[!ht]
\centering
\scalebox{0.58}{ 
\begin{tabular}{@{}lcccccc@{}}
\toprule
\textbf{Type of Poetry} & \multicolumn{2}{c}{Tang}              & \multicolumn{2}{c}{Song}              & \multicolumn{2}{c}{Yuan}              \\ \midrule
\textbf{Language}       & Chinese & \multicolumn{1}{l}{English} & Chinese & \multicolumn{1}{l}{English} & Chinese & \multicolumn{1}{l}{English} \\ \midrule
ChatGPT                 & 6.6     & 0.4                         & 4.4     & 0.6                         & 1.7     & 0.4                         \\
GPT4                    & 8.1     & 0.8                         & 7.3     & 0.9                         & 4.2     & 0.6                         \\ \bottomrule
\end{tabular}}
\caption{BLEU Scores from data validation experiments}
\label{data_validation}
\end{table}
\vspace{-0.6cm}



\subsection{Impact of Different Knowledge on Translation Performance}
\label{different_knowledge}



\begin{figure}[!th]
\centering
\includegraphics[width=0.8\linewidth]{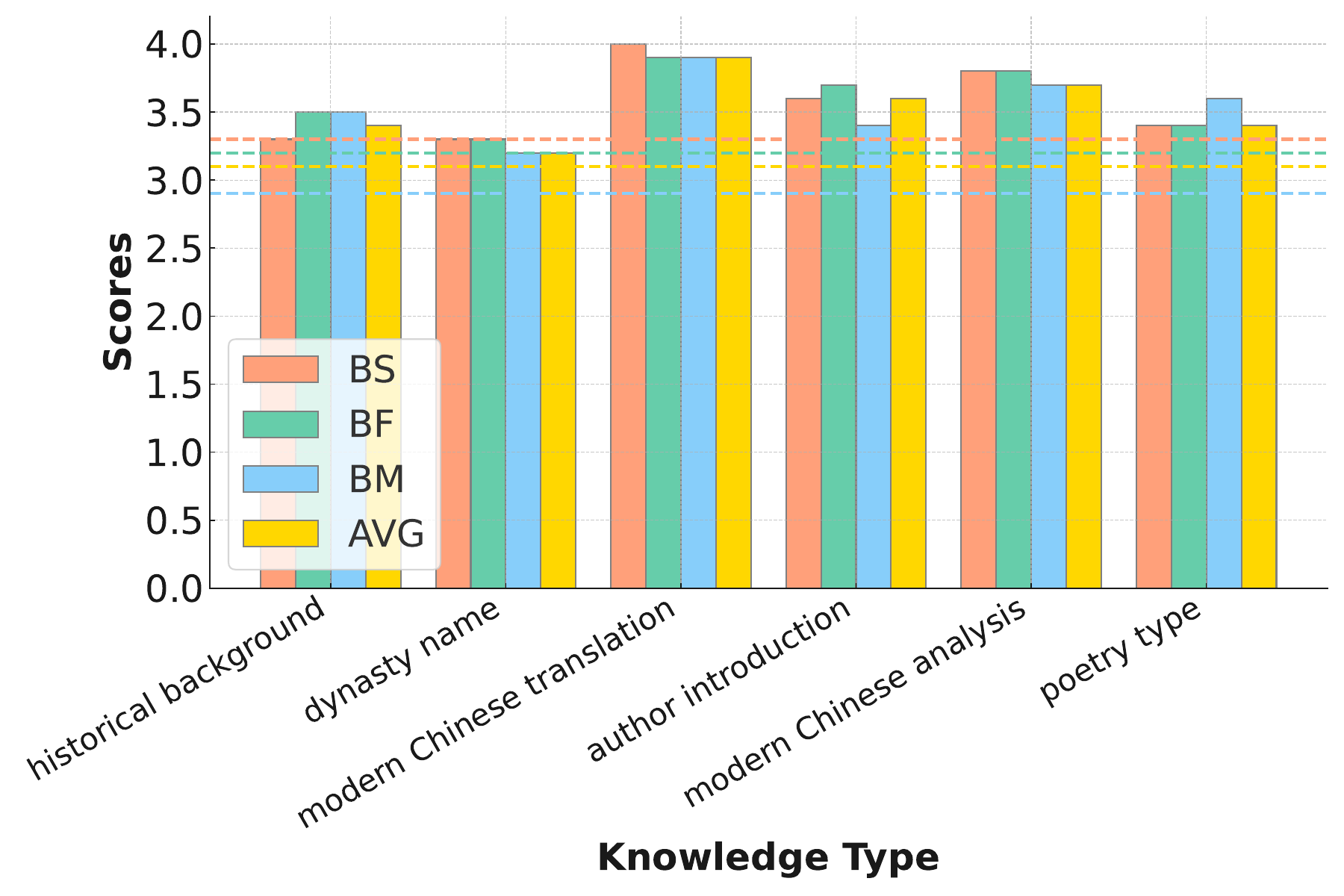} 
\caption{Experiment on the Impact of Different Knowledge of Classical Chinese Poetry on Translation. The \dashuline{dashed line} indicates not using knowledge, but directly translating the result through ChatGPT.}
\label{knowledge}
\end{figure}

The RAT method uses the Classical Chinese Poetry Knowledge Base for translation. To find out which type of knowledge is most helpful, this experiment adjusted the RAT method by using only one type of knowledge at a time and removing the Voter module (Figure \ref{framework}). The results (Figure \ref{knowledge}) showed that retrieval-based methods improved overall performance, proving that knowledge is important in translating Classical Chinese Poetry. Among the knowledge types, modern Chinese translation knowledge helped the most, suggesting that differences in style, meaning, and rhythm between Classical and modern Chinese make direct translation challenging.

\subsection{Ablation Study on Modern Chinese Translations in RAT Framework}

In Section \ref{different_knowledge}, the Modern Chinese translations in the Knowledge Base within the RAT framework show the greatest impact on output quality. To determine whether the improved results are solely due to these Modern Chinese translations, we conducted an ablation experiment and a case study to analyze this question in greater depth.

\begin{table}[!ht]
\centering
\scalebox{0.440}{ 
\begin{tabular}{@{}lccccccc@{}}
\toprule
                                                                & COMET $\uparrow$ & BLEURT $\uparrow$ & LLM-BM $\uparrow$ & LLM-BS $\uparrow$ & LLM-BF $\uparrow$ & LLM-Avg $\uparrow$ & BLEU-1 $\uparrow$ \\ \midrule
ChatGPT-RAT                                                     & 61.1  & 42.4   & 3.3    & 3.2    & 2.9    & 3.1     & 9.8  \\
\begin{tabular}[c]{@{}l@{}}ChatGPT-RAT\\ (only MC)\end{tabular} & 57.2  & 38.1   & 3.1    & 2.6    & 2.7    & 2.8     & 7.5  \\ \midrule
Vicuna-RAT                                                      & 60.1  & 26.9   & 3.0    & 2.5    & 3.3    & 2.9     & 17.6  \\
\begin{tabular}[c]{@{}l@{}}Vicuna-RAT\\ (only MC)\end{tabular}  & 53.1  & 26.9   & 2.7    & 2.4    & 2.5    & 2.5     & 19.0  \\ \bottomrule
\end{tabular}}
\caption{Ablation study results comparing the RAT framework with and without Modern Chinese (\textbf{MC}) translations in the Knowledge Base. }
\label{mc_ablition}
\end{table}

In Table \ref{mc_ablition}, the experimental results demonstrate that while Modern Chinese positively influences translation, the RAT method based on multi-knowledge still outperforms it. To further illustrate the limitations of Modern Chinese-based translation, we present three examples in Table \ref{tab:translations_case_mc} (additional two examples in the Appendix \ref{mc_ablation}).

\begin{table}[!ht]
\centering
\renewcommand\arraystretch{1.9}
\resizebox{0.9\linewidth}{!}{
\begin{tabular}{p{20cm}}
\toprule[2pt]
\fontsize{18}{16}\selectfont 
\textbf{Source:}  \begin{CJK*}{UTF8}{gbsn}红豆生南国，春来发几枝？愿君多采撷，此物最相思\end{CJK*} \\
\fontsize{18}{16}\selectfont 
\textbf{RAT:} Red beans grow in the south, sprouting many branches in spring. Pick them often, as they hold deep feelings of longing. \\

\fontsize{18}{16}\selectfont 
\textbf{RAT-only Modern Chinese:} Red beans grow in the sunny south, sprouting countless new branches every spring. I hope those who are missed will pick more of them, as they best express longing and love. \\

\fontsize{18}{16}\selectfont 
\textbf{Reference:} Red beans grow in the southern land, In spring, how many branches sprout? I wish you would gather them often, For they most evoke longing thoughts. \\

\bottomrule[2pt]
\end{tabular}}
\caption{Comparison of RAT, RAT-only Modern Chinese, and Reference Translations.}
\label{tab:translations_case_mc}
\end{table}
\vspace{-0.4cm}

The results of the case further support our viewpoint. Translations based on Modern Chinese tend to align more closely with general-domain sentence or passage-level translations and lack the adequacy, fluency, and elegance characteristics.

\subsection{Translation Challenges Across Different Types of Classical Chinese Poetry}

\begin{figure}[h]
\centering
\includegraphics[width=0.9\linewidth]{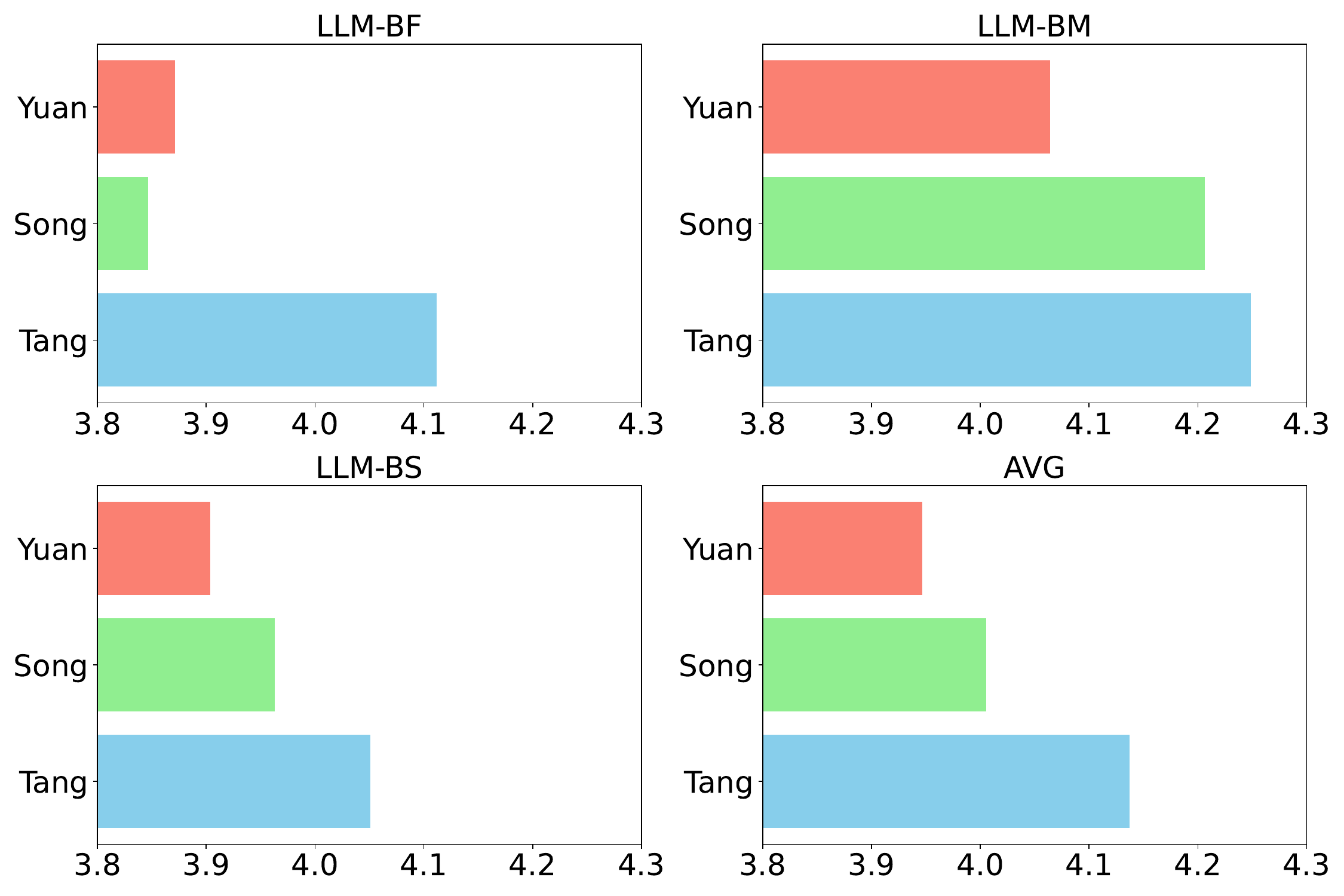} 
\caption{Experiment on the Impact of Different Types of Classical Chinese Poetry on Translation}
\label{type}
\end{figure}

To analyze the translation difficulty of different types of Classical Chinese poems (Tang, Song, Yuan) from a set of 608 poems, we used the RAT method and evaluated the results with LLM-BF, LLM-BM, LLM-BS, and LLM-AVG (Figure \ref{type}).

The results show consistent trends across types. Tang poetry is relatively easier to translate due to its stricter structure and brevity. Lower LLM-BF and LLM-BS scores highlight the challenge of translating poetic structure and rhythm. Higher LLM-BM scores indicate that retrieval-based methods produce more elegant translations.



\subsection{Human-centered Error Analysis}


To evaluate the RAT method's effectiveness and limitations, we manually assessed 50 randomly selected poems from the 608 test samples. Using a ChatGPT-based RAT method, translations were scored on a 1-5 scale for semantic adequacy, fluency, and elegance (Figures \ref{fig:poetry-evaluation-bs}, \ref{fig:poetry-evaluation_bf}, \ref{fig:poetry-evaluation_bm}). As shown in Table \ref{tab:exp1_results_human}, the results indicate a low proportion of Excellent (5-4) translations and a high proportion of Failed (2-1) ones. This highlights the challenges in PoetMT and the need to further refine the RAT method.

\begin{table}[!ht]
    \centering
    \scalebox{0.6}{ 
    \begin{tabular}{>{\centering\arraybackslash}m{3cm} >{\centering\arraybackslash}m{4cm} >{\centering\arraybackslash}m{3cm}}
        \toprule
        \textbf{Categories} & \textbf{Number of Sentences} & \textbf{Rate} \\
        \midrule
        \colorbox{green!20}{Excellent}           & 5                           & 10\%          \\
        \colorbox{yellow!20}{Decent}              & 23                          & 46\%          \\
        \colorbox{red!20}{Failed}              & 22                           & 44\%          \\
        \bottomrule
    \end{tabular}}
     \caption{Manual evaluation results of 50 RAT translations, categorized by performance.}
         \label{tab:exp1_results_human}
\end{table}
\vspace{-0.3cm}

Based on the results in Table \ref{tab:exp1_results_human}, we manually categorized the failed outcomes and provided case examples for clearer illustration in Table \ref{tab:translation_errors_case}.

\begin{table}[h!]
    \centering
    \scalebox{0.47}{
    \renewcommand{\arraystretch}{1.5}
    \begin{tabular}{>{\centering\arraybackslash}m{4cm} >{\centering\arraybackslash}m{2cm} m{8cm}}
        \toprule
        \textbf{Categories} & \textbf{Rate} & \textbf{Examples: Source/Error Result/Reference} \\
        \midrule
        Errors in handling polysemous words & 2/22 & \textit{Source:} \begin{CJK*}{UTF8}{gbsn}万壑树参天\end{CJK*}  
        \newline \textit{Error:} The trees in your valley scrape the sky  
        \newline \textit{Right:} In myriad gorges, trees touch the sky \\
        \midrule
        Lack of cultural context & 7/22 & \textit{Source:} \begin{CJK*}{UTF8}{gbsn}秦时明月汉时关\end{CJK*}  
        \newline \textit{Error:} The moon still shines on mountain passes as of yore  
        \newline \textit{Right:} Under the Qin moon, by the Han frontier \\
        \midrule
        Confusion in long sentence structures & 6/22 & \textit{Source:} \begin{CJK*}{UTF8}{gbsn}子弟每是个茅草岗沙土窝初生的兔羔儿乍向围场上走\end{CJK*}  
        \newline \textit{Error:} The young gallants are new-born bucks in chase of bunny  
        \newline \textit{Right:} Young ones are like rabbits, new to the hunt, Born in a thatch of grass, on sandy ground \\
        \midrule
        Incorrect translation of low-frequency vocabulary & 7/22 & \textit{Source:} \begin{CJK*}{UTF8}{gbsn}缚虎手\end{CJK*}  
        \newline \textit{Error:} Binding a tiger with bare hands  
        \newline \textit{Right:} Barehanded tiger fighting \\
        \bottomrule
    \end{tabular}}
        \caption{Translation Error Types with Examples.}
    \label{tab:translation_errors_case}
\end{table}

\section{Conclusion}
Our research reveals the challenges LLMs face in translating classical Chinese poetry, especially in cultural knowledge, fluency, and elegance. We introduced a GPT-4-based evaluation metric, demonstrating that existing large language models fall short in this task. Additionally, we introduced the RAT method to improve translation quality. This study is the first to evaluate the limitations of large language models in the translation of classical poetry It aims to inspire discussion within the translation community about the potential and future development of LLMs.

\section{Limitations}
The inherent challenges of translating classical poetry, such as the preservation of rhyme, tone, and aesthetic qualities, remain complex and subjective. Although the proposed GPT-4-based automatic evaluation metric has demonstrated consistency with human evaluation, these subjective dimensions still pose a significant challenge. 

\bibliography{custom}

\clearpage

\appendix

\section{Human Evaluations}

\subsection{Human Evaluation for ACC}
\label{sec:human_evaluation}
In this section, we conduct a human evaluation to measure translation quality.
We evaluated the adequacy of the translation. Four native English speakers were invited to participate. In the sentence-level adequacy task, the four experts scored each sentence for adequacy against the reference, awarding 1 point for fully adequate and 0 points for inadequate.

\subsection{Human Evaluation for BM score}
\label{human_evaluation_bm}

We conducted a human evaluation to measure the BM score, inviting four native English speakers to participate. A sample of 50 sentences was selected from the 758 data points, and the evaluators scored the translations based on the criteria in Figure \ref{fig:poetry-evaluation_bm}.

\section{Detail Prompt}

\subsection{Detailed prompt for Selector}
\label{detail stage1}

\begin{mdframed}[backgroundcolor=purple!10, linecolor=white, linewidth=2pt, roundcorner=10pt]\small
\textbf{Part-1: Selector:} Please identify the knowledge related to the content to translating this classical Chinese poem \{text\} from the \{rag context\} knowledge base.

\textbf{Input Text}: \begin{mdframed}[backgroundcolor=blue!10, linecolor=purple!10, linewidth=2pt, roundcorner=10pt]Source Poem, Sentence Length and Retrieved knowledge\end{mdframed}\small

\textbf{Output Text}: \begin{mdframed}[backgroundcolor=yellow!10, linecolor=purple!10, linewidth=2pt, roundcorner=10pt] Refined knowledge.\end{mdframed}\small
\end{mdframed}

\subsection{Detailed prompt for Translator}
\label{detail stage2}

\begin{mdframed}[backgroundcolor=purple!10, linecolor=white, linewidth=2pt, roundcorner=10pt]\small
\textbf{Part-2: Translator:} Please translate this classial a Chinese poem \{translate type\} into a English poem \{translate type\}: Explanation:\{rag context\} Poem:\{text\}

\textbf{Input Text}: \begin{mdframed}[backgroundcolor=blue!10, linecolor=purple!10, linewidth=2pt, roundcorner=10pt]Source Poem, Retrieved knowledge and Potery Type\end{mdframed}\small

\textbf{Output Text}: \begin{mdframed}[backgroundcolor=yellow!10, linecolor=purple!10, linewidth=2pt, roundcorner=10pt] Translated English Poem\end{mdframed}\small
\end{mdframed}

\newpage

\subsection{Detailed prompt for Voter}
\label{detail stage3}

\begin{mdframed}[backgroundcolor=purple!10, linecolor=white, linewidth=2pt, roundcorner=10pt]\small
\textbf{Part-3: Iterative Refinement:} Using the classical Chinese poem \{src\_text\} as a source, compare six translation candidates to determine the highest quality result. Avoid including unrelated content. Here are the candidates: First, \{s1\}; second, \{s2\}; third, \{s3\}; fourth, \{s4\}; fifth, \{s5\}; sixth, \{s6\}.

\textbf{Input Text}: \begin{mdframed}[backgroundcolor=blue!10, linecolor=purple!10, linewidth=2pt, roundcorner=10pt]Source Sentence, Translated Resluts based on six knowledge\end{mdframed}\small

\textbf{Output Text}: \begin{mdframed}[backgroundcolor=yellow!10, linecolor=purple!10, linewidth=2pt, roundcorner=10pt] Translated Result
\end{mdframed}\small
\end{mdframed}

\subsection{Detailed prompt for Extractor}
\label{detail stage4}

\begin{mdframed}[backgroundcolor=purple!10, linecolor=white, linewidth=2pt, roundcorner=10pt]\small
\textbf{Part-4:Understanding-Based Translation:} Extract only translation-relevant content from \{target text\} based on \{text\}.
\textbf{Input Text}: \begin{mdframed}[backgroundcolor=blue!10, linecolor=purple!10, linewidth=2pt, roundcorner=10pt]
The final translation result.
\end{mdframed}\small

\textbf{Output Text}: \begin{mdframed}[backgroundcolor=yellow!10, linecolor=purple!10, linewidth=2pt, roundcorner=10pt]Target Sentence $t$\end{mdframed}\small
\end{mdframed}

\subsection{Comparative Methods}
\label{sec:comparative_methods}

The following content will provide detailed descriptions of these comparative methods: 
\begin{itemize}

\item \textbf{Baseline}, standard zero-shot translation is performed in ChatGPT \cite{ouyang2022training} and GPT-4 \cite{achiam2023gpt}. The temperature parameter set to 0, which is the default value for our experiments.

\item \textbf{5-Shot} \cite{hendy2023good}, involves prepending five high-quality labelled examples from the training data to the test input.

\item \textbf{Rerank} \cite{DBLP:conf/eamt/MoslemHKW23} was conducted with the identical prompt as the baseline, employing a temperature of 0.3 \citep{moslem2023adaptive}. Three random samples were generated and combined with the baseline to yield four candidates. The optimal candidate was chosen through GPT4.

\item \textbf{Refine \cite{chen2023iterative}} first requests a translation from ChatGPT, then provides the source text and translation results, and obtains a refined translation through multiple rounds of modifications by mimicking the human correction process.


\item \textbf{MAD \cite{liang2023encouraging}} enhance the capabilities of LLMs by encouraging divergent thinking. In this method, multiple agents engage in a debate, while an agent oversees the process to derive a final solution.

\item \textbf{EAPMT} \cite{Wang2024WhatIT} leverages the explanation of monolingual poetry as guidance information to
achieve high-quality translations from Chinese poetry to English poetry.

\item \textbf{Dual-Reflect\cite{chen-etal-2024-dual}} provide supervisory signals for large models to reflect on translation results through dual learning, thereby iteratively improving translation performance (the maximum number of iterations is set to 5).

\item \textbf{RAT} is the proposed method in this work.


\end{itemize}

\subsection{Detailed prompt for Beauty of Sound}
\label{bs_score_appendix}

For evaluation of the beauty of form, the detailed prompt is displayed in Figure \ref{fig:poetry-evaluation_bf}

\begin{figure}[h!]
    \begin{mdframed}[backgroundcolor=chinese_red, linecolor=chinese_red, linewidth=2pt, roundcorner=10pt]
    \small
    \textbf{/* Task prompt */} 
    
    Evaluate the beauty of sound in the given  Chinese translation of classical poetry. Focus on whether the  translation achieves harmonious sound, adherence to strict metrical rules, and a rhythm 
    
    1 point: Poor translation, lacks harmony and adherence to metrical rules, and fails to capture the beauty of sound.
    
    2 point: Below average, some rhyme and meter present but with noticeable imperfections and awkwardness.
    
    3 point: Basic translation, captures some aspects of sound beauty but
    with several imperfections in rhyme, meter, or rhythm.
    
    4 point: Good translation, mostly harmonious with minor imperfections 
    in sound quality or adherence to metrical rules.
    
    5 point: Excellent translation, achieves harmonious sound, precise wording, strict adherence to metrical rules, and a smooth, dynamic rhythm.
    
    \textbf{/* Input Data */}: 
    \begin{mdframed}[backgroundcolor=chinese_red_small, linecolor=chinese_red_small, linewidth=2pt, roundcorner=10pt]
    Original Chinese poem: \{source\}
    
    English translation: \{translation\}
    
    Evaluation (score only):
    \end{mdframed}
    
    \textbf{/*Output Text */}: 
    \begin{mdframed}[backgroundcolor=chinese_red_small, linecolor=chinese_red_small, linewidth=2pt, roundcorner=10pt] 
    \{score\} 
    \end{mdframed}
    \end{mdframed}
    \caption{Evaluation of the beauty of sound in Chinese translation of classical poetry}
    \label{fig:poetry-evaluation-bs}
\end{figure}

\subsection{Detailed prompt for Beauty of Form}
\label{bf_score}

For evaluation of the beauty of form, the detailed prompt is displayed in Figure \ref{fig:poetry-evaluation_bf}

\begin{figure}[h!]
    \begin{mdframed}[backgroundcolor=chinese_red, linecolor=chinese_red, linewidth=2pt, roundcorner=10pt]
    \small
    \textbf{/* Task prompt */} 
    
    Evaluate the translation of the given Chinese classical poem into English. Focus on whether the translation maintains consistency with the source poem's structure, including the alignment of line numbers and balanced phrasing.
    
    1 point: Poor translation, disregards the poem's structure, and fails to convey its aesthetic qualities.
    
    2 point: Some attempt to maintain structure but lack alignment and aesthetic consistency.
    
    3 point: Basic structural elements are maintained but with noticeable imperfections in alignment and phrasing.
    
    4 point: Good translation, with most structural elements preserved and minor issues in phrasing and alignment.
    
    5 point: Excellent translation, accurately preserving the structure, alignment, and aesthetic qualities of the original poem.
        
    \textbf{/* Input Data */}: 
    \begin{mdframed}[backgroundcolor=chinese_red_small, linecolor=chinese_red_small, linewidth=2pt, roundcorner=10pt]
    Original Chinese poem: \{source\}
    
    English translation: \{translation\}
    
    Evaluation (score only):
    \end{mdframed}
    
    \textbf{/*Output Text */}: 
    \begin{mdframed}[backgroundcolor=chinese_red_small, linecolor=chinese_red_small, linewidth=2pt, roundcorner=10pt] 
    \{score\} 
    \end{mdframed}
    \end{mdframed}
    \caption{Evaluation of the beauty of form in Chinese translation of classical poetry}
    \label{fig:poetry-evaluation_bf}
\end{figure}

\subsection{Detailed prompt for Beauty of Meaning}
\label{bm_score}

For evaluation of the beauty of meaning, the detailed prompt is displayed in Figure \ref{fig:poetry-evaluation_bm}

\begin{figure}[h!]
    \begin{mdframed}[backgroundcolor=chinese_red, linecolor=chinese_red, linewidth=2pt, roundcorner=10pt]
    \small
    \textbf{/* Task prompt */} 
    
    Evaluate the translation of Chinese classical poetry for the beauty of meaning, focusing on whether the translation effectively conveys the themes, emotions, and messages of the original. This includes the use of concise and precise language to create vivid imagery and a rich atmosphere.
    
    1 point: Poor translation, fails to convey the depth and richness of the original poetry.
    
    2 point: Basic translation with significant shortcomings in capturing themes, emotions, and messages.
    
    3 point: Satisfactory translation, conveys basic themes and emotions but lacks refinement or depth.
    
    4 point: Good translation, effectively captures most themes, emotions, and messages with minor imperfections.
    
    5 point: Excellent translation, accurately conveys the depth, richness, and atmosphere of the original poetry with full thematic and emotional resonance.

    \textbf{/* Input Data */}: 
    \begin{mdframed}[backgroundcolor=chinese_red_small, linecolor=chinese_red_small, linewidth=2pt, roundcorner=10pt]
    Original Chinese poem: \{source\}
    
    English translation: \{translation\}
    
    Evaluation (score only):
    \end{mdframed}
    
    \textbf{/*Output Text */}: 
    \begin{mdframed}[backgroundcolor=chinese_red_small, linecolor=chinese_red_small, linewidth=2pt, roundcorner=10pt] 
    \{score\} 
    \end{mdframed}
    \end{mdframed}
    \caption{Evaluation of the beauty of meaning in Chinese translation of classical poetry}
    \label{fig:poetry-evaluation_bm}
\end{figure}

\section{Supplementary Experiment}

\subsection{LLM-based Metric Consistency}
\begin{figure}[!ht]
\centering
\includegraphics[width=\linewidth]{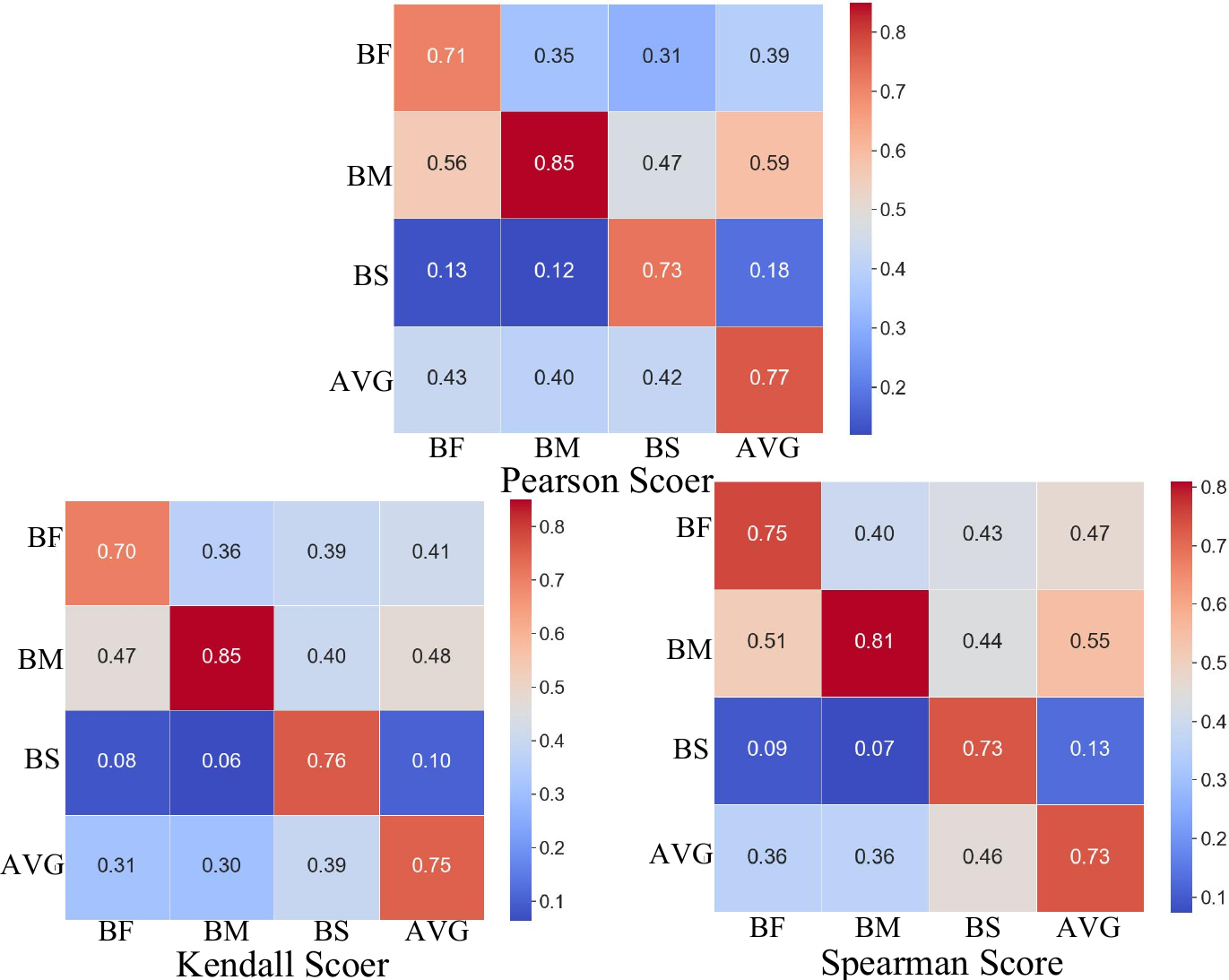} 
\caption{\textbf{LLM-based Metric Consistency Experiment.} In the heatmap, the horizontal axis represents the human evaluation results, and the vertical axis represents the LLM evaluation results.}
\label{corr_figure}
\end{figure}

This experiment evaluated whether the proposed LLM-based metrics (LLM-BS, LLM-BF, LLM-BM and LLM-AVG) accurately reflect Beauty of Sound, Beauty of Form, Beauty of Meaning, and overall translation quality. We conducted pairwise correlation tests between human and LLM-based evaluations using Pearson, Spearman, and Kendall correlation coefficients. The results are shown in Figure \ref{corr_figure}.

The experimental results indicate that, among all correlation coefficients, the consistency results based on the same annotations are significantly higher than the other results. This demonstrates the rationality of the evaluation settings for LLM-BS, LLM-B, LLM-BM, and LLM-AVG in the experiment.

\subsection{Additional Evaluation of Adequacy of Open-source LLMs}
\label{adequcay_of_llm}


\begin{table}[!ht]\centering
\scalebox{0.8}{ 
\begin{tabular}{@{}lccc@{}}
\toprule
Methods                            & LLM-BM       & \multicolumn{1}{l}{Human-BM} & ACC           \\ \midrule
\textbf{Llama3-8B}                   &              &                                      &               \\
\quad+Zero-Shot    & 2.1          & 1.1                                  & 32.5          \\
\quad+Rerank       & 2.1          & 1.2                                  & 32.4          \\
\quad+Dual-Reflect & 2.5          & 1.7                                  & 34.4          \\
\quad+MAD          & 2.6          & 1.9                                  & 37.3          \\
\quad+RAT          & \textbf{2.9} & \textbf{2.4}                         & \textbf{59.9} \\ \midrule
\textbf{Qwen-72B}                   &              &                                      &               \\
\quad+Zero-Shot    & 3.1          & 2.2                                  & 43.9          \\
\quad+Rerank       & 3.3          & 2.3                                  & 42.7         \\
\quad+Dual-Reflect & 3.0          & 2.0                                  & 46.3          \\
\quad+MAD & 3.1          & 2.4                                  & 47.5          \\
\quad+RAT          & \textbf{3.3} & \textbf{2.8}                         & \textbf{55.4} \\
\bottomrule
\end{tabular}}
\caption{The LLM-BM and human-annotated results of the Adequacy in Sentence-Level PoetMT benchmark Translation.}
\label{human_main}
\end{table}

\subsection{Additional Examples of Modern Chinese-Based Translation Limitations}
\label{mc_ablation}

To provide a more comprehensive understanding of the limitations inherent in Modern Chinese-based translation, Table \ref{tab:translations_case_mc_appendix} presents two additional examples. These examples further illustrate scenarios where the RAT method demonstrates superior performance due to its multi-knowledge integration, highlighting the gaps left by relying solely on Modern Chinese interpretations.

\begin{table*}[!ht]
\centering
\renewcommand\arraystretch{1.8} 
\resizebox{\textwidth}{!}{
\begin{tabular}{p{20cm}}
\toprule[2pt]

\textbf{Source:} \begin{CJK*}{UTF8}{gbsn}水纹珍簟思悠悠，千里佳期一夕休。从此无心爱良夜，任他明月下西楼。\end{CJK*} \\

\textbf{RAT:} Ripples stir the jade mat as thoughts drift endlessly afar,  
A thousand miles of vows dissolve beneath the evening star.  
No longer shall fine nights find a place within my heart,  
Let the moonlight wane and vanish, where its path departs. \\

\textbf{RAT-only Modern Chinese:} Lying on the precious bamboo mat, my thoughts drift away with the flowing water.  
Suddenly, the winds and clouds have changed, and the cherished promise spanning a thousand miles has been shattered in an instant.  
From now on, I no longer have the heart to appreciate the beautiful moments,  
whether the moon rises above the eastern tower or sets beneath the western tower. \\

\textbf{Reference:} Ripples on the jade mat, thoughts drift far and wide,  
A thousand miles of promise, in one night, all denied.  
From now on, I'll have no heart for such fine nights,  
Let the bright moon set where it will, beyond my sight. \\

\midrule[1pt]

\textbf{Source:} \begin{CJK*}{UTF8}{gbsn}林暗草惊风，将军夜引弓。平明寻白羽，没在石棱中。\end{CJK*} \\

\textbf{RAT:} Through shadowed woods and rustling grass,  
The general's bow is drawn at last.  
By dawn he seeks the arrow's flight,  
Its white fletch lodged in stone, out of sight. \\

\textbf{RAT-only Modern Chinese:} In the dim woods, the grass suddenly rustles in the wind, prompting the general to quickly draw his bow and shoot in the cover of night.  
At dawn, he searches for the white-fletched arrow fired the previous evening, finding its tip deeply embedded in a massive stone. \\

\textbf{Reference:} Amid the dark woods and startled grasses,  
The general draws his bow at night.  
At dawn he seeks the white fletching,  
Embedded in the stone's edge, out of sight. \\

\bottomrule[2pt]
\end{tabular}}
\caption{Comparison of RAT, RAT-only Modern Chinese, and Reference Translations.}
\label{tab:translations_case_mc_appendix}
\end{table*}

\subsection{Knowledge Base Case}
\label{base_case_section}

A case regarding the Classical Chinese Poetry Knowledge Base is shown in Table \ref{base case}, demonstrating relevant knowledge of classical Chinese poetry.

\begin{table*}[!htp]
\centering
\resizebox{\linewidth}{!}{
\renewcommand\arraystretch{1.2}
\begin{tabular}{p{20cm}}
\toprule[2pt]

\textbf{Classical Chinese Poem: }\begin{CJK*}{UTF8}{gbsn}慈母手中线，游子身上衣。临行密密缝，意恐迟迟归。谁言寸草心，报得三春晖。\end{CJK*} \textcolor{gray!110}{(From the threads a mother's hand weaves, A gown for parting son is made.Sown stitch by stitch before he leaves, For fear his return be delayed. Such kindness as young grass receives from the warm sun can't be repaid.)}\\
\textbf{Historical Background:}  \begin{CJK*}{UTF8}{gbsn}《游子吟》写在溧阳。孟郊早年漂泊无依，一生贫困潦倒，直到五十岁时才得到了一个溧阳县尉的卑微之职，结束了长年的漂泊流离生活，便将母亲接来住。诗人仕途失意，饱尝了世态炎凉，此时愈觉亲情之可贵，于是写出这首发于肺腑，感人至深的颂母之诗。\end{CJK*} \textcolor{gray!110}{("Song of the Parting Son" was written in Liyang. In his early years, Meng Jiao lived a wandering and destitute life, experiencing poverty throughout his existence. It was not until he was fifty that he obtained a modest position as a county official in Liyang, which finally ended his years of wandering. He then brought his mother to live with him. Having faced the disappointments of his career and the coldness of society, he grew increasingly aware of the preciousness of familial bonds. Thus, he composed this deeply heartfelt poem in honour of his mother.)}  \\
\textbf{Dynasty Name:} \begin{CJK*}{UTF8}{gbsn}唐代\end{CJK*} \textcolor{gray!110}{(Tang Dynasty)}       \\
\textbf{Morden Chinese Translation:} \begin{CJK*}{UTF8}{gbsn}慈母用手中的针线，为远行的儿子赶制身上的衣衫。临行前一针针密密地缝缀，怕的是儿子回来得晚衣服破损。有谁敢说，子女像小草那样微弱的孝心，能够报答得了像春晖普泽的慈母恩情呢？\end{CJK*} \textcolor{gray!110}{(A loving mother uses her needle and thread to make clothes for her son, who is about to embark on a journey. She stitches each seam tightly, fearing that her son may return late and the clothes will be worn out. Who can dare say that a child's feeble filial piety, like a small blade of grass, can repay the boundless kindness of a mother, akin to the nurturing warmth of spring sunlight?)}   \\
\textbf{Author Introduction:} \begin{CJK*}{UTF8}{gbsn}孟郊，(751-814)，唐代诗人。字东野。汉族，湖州武康（今浙江德清）人，祖籍平昌（今山东临邑东北），先世居洛阳（今属河南）。唐代著名诗人。现存诗歌500多首，以短篇的五言古诗最多，代表作有《游子吟》。有"诗囚"之称，又与贾岛齐名，人称"郊寒岛瘦"。元和九年，在阌乡(今河南灵宝)因病去世。张籍私谥为贞曜先生。\end{CJK*} \textcolor{gray!110}{(Meng Jiao (751-814) was a poet of the Tang Dynasty. His courtesy name was Dongye. He was of Han ethnicity and hailed from Wukang, Huzhou (present-day Deqing, Zhejiang), with ancestral roots in Pingchang (northeast of present-day Linyi, Shandong). His family originally resided in Luoyang (now in Henan). A renowned poet of the Tang era, he has over 500 surviving poems, most of which are short five-character ancient verses. His notable works include "Song of the Parting Son." He was known as the "Poet Prisoner" and was contemporaneous with Jia Dao, with the phrase "Jiao Han, Dao Shou" used to describe them together. He passed away in the ninth year of the Yuanhe era, in Wanquan (present-day Lingbao, Henan), due to illness. Zhang Ji posthumously honoured him with the title of "Mr Zhenyao.")}\\
\textbf{Modern Chinese Analysis: } \begin{CJK*}{UTF8}{gbsn}开头两句用"线"与"衣"两件极常见的东西将"慈母"与"游子"紧紧联系在一起，写出母子相依为命的骨肉感情。三、四句通过慈母为游子赶 制出门衣服的动作和心理的刻画，深化这种骨肉之情。母亲千针万线"密密缝"是因为怕儿子"迟迟"难归。前面四句采用白描手法，不作任何修饰，但慈母的形象真切感人。最后两句是作者直抒胸臆，对母爱作尽情的讴歌。这两句采用传统的比兴手法：儿女像区区小草，母爱如春天阳光。\end{CJK*} \textcolor{gray!110}{(The opening two lines connect "the loving mother" and "the wandering son" through the commonplace items of "thread" and "clothes," highlighting the deep bond of flesh and blood between them. In the third and fourth lines, the mother's actions and thoughts as she makes clothes for her son further deepen this familial affection. The mother's meticulous stitching is driven by her fear that her son will return late. The first four lines employ a straightforward style, without embellishment, yet the image of the loving mother is vivid and touching. The final two lines express the author's heartfelt emotions, celebrating maternal love. These lines use traditional metaphorical techniques: children are like fragile blades of grass, while maternal love resembles the warm sunlight of spring.)}\\
\textbf{Poetry Type: } \begin{CJK*}{UTF8}{gbsn} 唐诗三百首,乐府,赞颂,母爱\end{CJK*} \textcolor{gray!110}{(Three Hundred Tang Poems, Yuefu, Panegyric, Maternal Love.)}\\
\bottomrule[2pt]
\end{tabular}}
\caption{A case about Classical Chinese Poetry Knowledge Base.}
\label{base case}
\end{table*}

\end{document}